\documentclass{ifacconf}

\usepackage{color}
\usepackage{bm}

\usepackage{graphicx}
\usepackage[font={small},width=.47\textwidth]{caption}
\usepackage{amsmath}
\usepackage{amssymb}

\usepackage{nicefrac} 

\usepackage{tikz}
\usetikzlibrary{arrows} 
\usetikzlibrary{calc} 
\usetikzlibrary{plotmarks}
\usepackage{pgfplots}

\usepackage[capitalize]{cleveref} 

\usepackage{comment}

\newtheorem{remark}{Remark}

\usepackage{graphicx}      
\usepackage{natbib}        
\begin{document}
\begin{frontmatter}

\title{Optimization-Based On-Road Path Planning for Articulated Vehicles\thanksref{footnoteinfo}} 

\thanks[footnoteinfo]{This work was partially supported by the Wallenberg AI, Autonomous Systems and Software Program (WASP) funded by the Knut and Alice Wallenberg Foundation.
\newline
This work has been accepted to IFAC for publication under a Creative Commons Licence CC-BY-NC-ND.}

\author[First]{Rui Oliveira} 
\author[Second]{Oskar Ljungqvist}
\author[Third]{Pedro F. Lima}
\author[First]{Bo Wahlberg}

\address[First]{Division of Decision and Control Systems, KTH Royal Institute of Technology, Stockholm, Sweden (e-mail: \{rfoli, bo\}@kth.se).}
\address[Second]{Division of Automatic Control, Link\"oping University, Link\"oping, Sweden (e-mail: oskar.ljungqvist@liu.se).}
\address[Third]{Autonomous Transport Solutions, Scania CV AB, S\"odert\"alje, Sweden (e-mail: pedro.lima@scania.com).}

\begin{abstract}                
Maneuvering an articulated vehicle on narrow road stretches is often a challenging task for a human driver. Unless the vehicle is accurately steered, parts of the vehicle's bodies may exceed its assigned drive lane, resulting in an increased risk of collision with surrounding traffic. In this work, an optimization-based path-planning algorithm is proposed targeting on-road driving scenarios for articulated vehicles composed of a tractor and a trailer. To this end, we model the tractor-trailer vehicle in a road-aligned coordinate frame suited for on-road planning. Based on driving heuristics, a set of different optimization objectives is proposed, with the overall goal of designing a path planner that computes paths which minimize the off-track of the vehicle bodies swept area, while remaining on the road and avoiding collision with obstacles. The proposed optimization-based path-planning algorithm, together with the different optimization objectives, is evaluated and analyzed in simulations on a set of complicated and practically relevant on-road planning scenarios using the most challenging tractor-trailer dimensions.

\end{abstract}


\end{frontmatter}

\def\FigureCaptionTextSize{\normalsize}

\def\optObjCenterDrivingTractor{ J_{ \text{center,tractor} } }
\def\optObjCenterDrivingTrailer{ J_{ \text{center,trailer} } }
\def\optObjCenterCrossTerm{ J_{ \text{tractor} - \text{trailer} } }
\def\optObjSmooth{ J_{ \text{smooth} } }

\def\Veh{\text{veh}}
\def\Tra{\text{tra}}
\def\Hit{\text{hit}}
\def\DVeh{L_1}
\def\DTra{L_2}
\def\DHit{M_1}

\def\EY{e_{y}}
\def\EPsi{e_{\psi}}
\def\BetaTwo{\beta_{1}}

\def\STrailer{s_{\Tra}}
\def\EYTrailer{e_{y,\Tra}}
\def\EPsiTrailer{e_{\psi,\Tra}}
\def\STrailerApprox{\hat{s}_{\Tra}}
\def\EYTrailerApprox{\hat{e}_{y,\Tra}}
\def\EPsiTrailerApprox{\hat{e}_{\psi,\Tra}}

\def\VectorEY{\mathbf{\EY}}
\def\VectorEPsi{\mathbf{\EPsi}}
\def\VectorBetaTwo{\mathbf{\BetaTwo}}
\def\VectorEYTrailer{\mathbf{\EYTrailer}}
\def\VectorEPsiTrailer{\mathbf{\EPsiTrailer}}
\def\VectorBetaTwoTrailer{\mathbf{\BetaTwoTrailer}}
\def\VectorU{\mathbf{u}}
\def\VectorK{\mathbf{\kappa}}

\def\SBar{\bar{s}}
\def\EYBar{\bar{e}_y}
\def\EPsiBar{\bar{e}_\psi}
\def\BetaTwoBar{\bar{\beta}_1}

\def\STrailerBar{\bar{s}_{\Tra}}
\def\EYTrailerBar{\bar{e}_{y,\Tra}}
\def\EPsiTrailerBar{\bar{e}_{\psi,\Tra}}

\def\DeltaApproxEY{\delta \EY}
\def\DeltaApproxEPsi{\delta \EPsi}
\def\DeltaApproxBetaTwo{\delta \BetaTwo}

\def\VehState{z}

\def\DiscretizationSamplingDistance{\Delta s}
\def\linearizationReferenceVariablesS{ \mathbf{\bar{s}} }
\def\linearizationReferenceVariablesEy{ \mathbf{\bar{e}_y} }
\def\linearizationReferenceVariablesEpsi{ \mathbf{\bar{e}_\psi} }
\def\linearizationReferenceVariablesBetaTwo{ \mathbf{\bar{\beta}_1} }
\def\linearizationReferenceVariablesU{ \mathbf{\bar{u}} }
\def\linearizationReferenceVariablesK{ \mathbf{\bar{\kappa}} }

\def\optObjCenterDriving{ J_{ \mathrm{center} } }
\def\optObjOverhang{ J_{ \mathrm{overhang} } }
\def\optObjSmooth{ J_{ \mathrm{smooth} } }

\def\optConstraintObsMatrix{ P_{i} }
\def\optConstraintObsOffset{ p_{i} }
\def\optConstraintObsPosition{ p_{e_y}^{\text{obs},i} }

\def\optConstraintMaxU{ u_{\max} }
\def\optConstraintMaxUDot{ {u}_{\max}' }
\def\optConstraintMaxUDotDot{ {u}_{\max}'' }
\def\optConstraintFirstDifferenceMatrix{ \mathbf{D_1} }
\def\optConstraintSecondDifferenceMatrix{ \mathbf{D_2} }

\def\optConstraintMaxK{ \kappa_{\max} }
\def\optConstraintMaxKDot{ {\kappa}_{\max}' }

\section{Introduction}

\begin{figure}[t!]
\centering
\includegraphics[width=0.98\columnwidth]{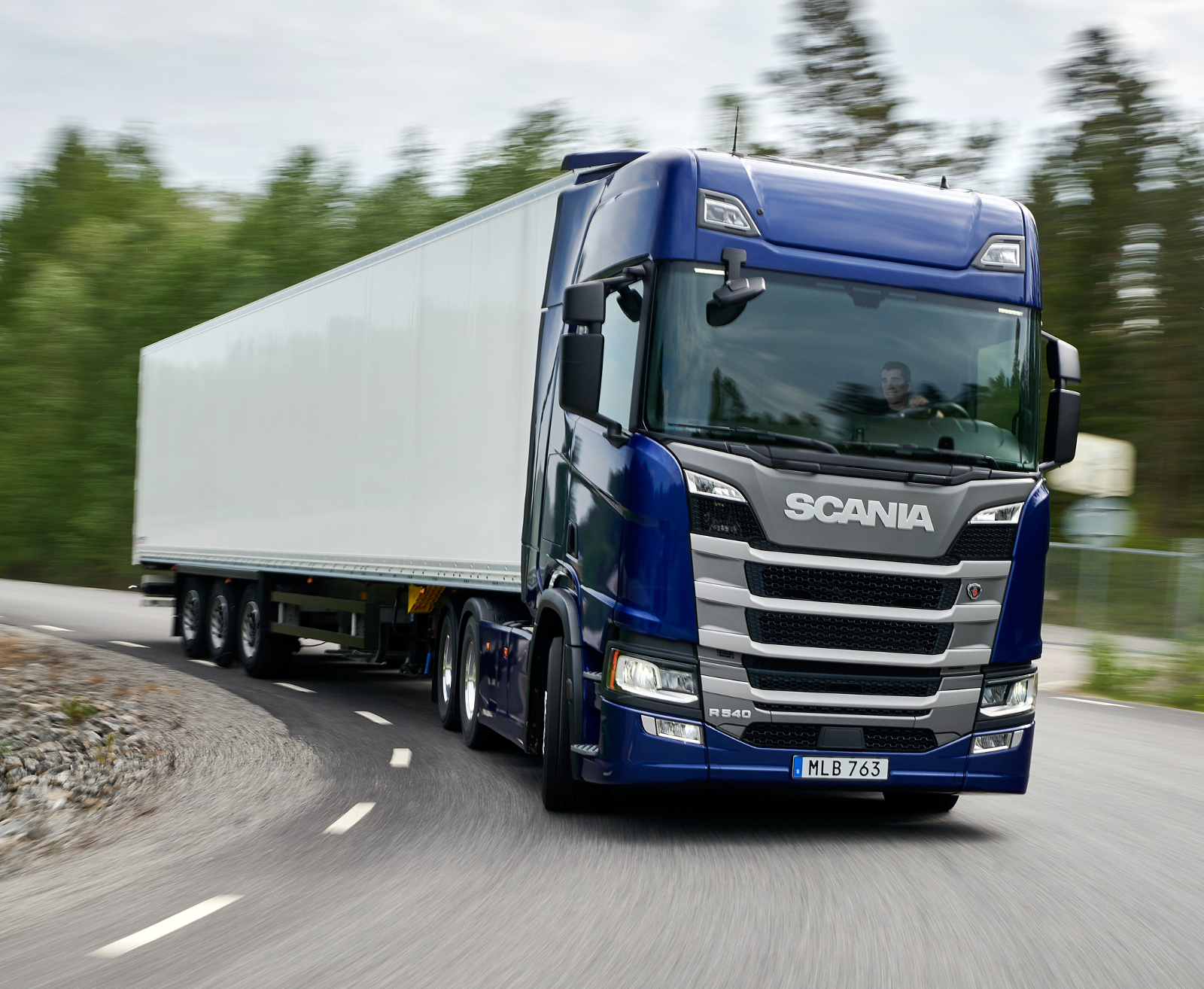}
\caption{{\FigureCaptionTextSize A tractor-trailer vehicle whose dimensions are used in this work. A tractor-trailer combination is an example of an articulated vehicle. This work considers such vehicle combinations, which also include articulated buses. (courtesy of Scania CV AB)}}
\label{fig:truck-trailer-photo}
\end{figure}

Autonomous driving technologies are expected to revolutionize the transportation industry by increasing the efficiency and safety of vehicles.
A significant part of vehicles in today's traffic is articulated, such as the tractor-trailer combination shown in~\cref{fig:truck-trailer-photo}, and are responsible for 9.2\% of all distance driven in roads nowadays~\citep{Bureau:2019:USVehicleMiles}.
Despite playing an essential role in the society, the trucking industry is currently facing a shortage of drivers~\citep{ATRI:2016:AutonomousDrivingImpactOnTrucking}.
Together, these factors motivate the development of self-driving technologies targeting tractor-trailer vehicles.

A tractor-trailer vehicle is characterized by two bodies of large dimensions, making it very difficult to successfully maneuver such vehicles on narrow road stretches, such as roundabouts or city streets. 
One difficulty arises from the so-called off-tracking effect that forces the trailer to take a short cut when the leading tractor performs a sharp turn~\citep{Jujnovich:2013:SteeringControl,altafini:2003OffTracking}.
As a consequence, the problem of computing optimized paths that take the off-tracking effect into account, while avoiding collision with surrounding obstacles, is both practically relevant and an important subject which requires the development of specialized solutions. 

In the literature, off-track reduction of articulated vehicles has so far mainly been treated as a path-following control problem~\citep{altafini:2002Following,Jujnovich:2013:SteeringControl,altafini:2003OffTracking,Liu:2018minimumSwept,michalek:2015motionOffTrack}. Even though these control solutions reduce the off-tracking effect, they disregard obstacle-avoidance constraints, as well as the dimensions of the vehicle bodies. In this work, we treat the off-track reduction problem as a path-planning problem in order to guarantee collision avoidance of all vehicle bodies as well as minimizing a general optimization objective. 

Path planning deals with the problem of computing paths that an autonomous vehicle can follow.
These paths must be collision-free and comply with the nonholonomic constraints imposed on the vehicle.
Additionally, they should also be optimized based on multiple criteria, such as comfort, safety, and efficiency.
Path planning is a widely studied subject with applications ranging from robotic manipulators to self-driving vehicles~\citep{LaValle:2006:Planning}. 

The research on path planning for tractor-trailer vehicles has mainly been focusing on off-road scenarios~\citep{Ljungqvist:2019:Path,Lamiraux:1999:hillary,evestedt:2016CLRRT,Li:2019trajectory}, semi-structured scenarios~\citep{Lamiraux:2005:A380,Beyersdorfer:2013tractortrailer}, or low curvature roads~\citep{van2015real}. Even though on-road path planning for passenger cars has received much attention~\citep{Katrakazas:2015:Survey, Frazzoli:2016:Survey}, only a limited amount of work considers this critical problem for articulated vehicles, which requires tailored solutions to comply with the large vehicle dimensions.

In this work, we consider the path-planning problem for tractor-trailer vehicles during on-road driving scenarios, characterized by narrow passages with sharp turns. To this end, the tractor-trailer vehicle is modeled in a so-called road-aligned coordinate frame, as is commonly used for on-road driving scenarios~\citep{Katrakazas:2015:Survey, Frazzoli:2016:Survey}.
The path planning problem is then formulated as a nonlinear optimization problem and solved using a Sequential Quadratic Programming (SQP) approach based on previous work in~\cite{Oliveira:2019:BusDriving}.
We present a set of novel optimization objectives to use within the optimization-based path planning algorithm.
These objectives are tailored to minimize the off-tracking effect of the swept area by the bodies of the tractor-trailer.
We evaluate the impact of the proposed objectives on the planned paths in a set of simulations, concerning metrics such as maximum off-track, road-centering alignment of the total swept area by the tractor-trailer vehicle, and computation time.
In summary, the contributions of this work are:
\begin{itemize}
\item proposal and evaluation of different optimization criteria suitable for on-road path planning of articulated vehicles;
\item implementation of a sequential quadratic programming (SQP) solver, which ensures smooth driving while guaranteeing precise obstacle avoidance;
\item a sequential method for computing the off-tracking, as well as approximate partial derivatives, of each point of the vehicle bodies suitable for numerical optimization approaches;
\item simulation results showing the proposed path planner's ability to solve complicated on-road planning scenarios while considering the most challenging tractor-trailer dimensions.
\end{itemize}

The remainder of the paper is organized as follows. The proposed road-aligned model of the tractor-trailer vehicle is presented in~\cref{sec:vehicle_model}. 
In~\cref{sec:problem_formulation}, the optimization-based path planner is presented, as well as a set of optimization objectives tailored for on-road driving for tractor-trailer vehicles.
Simulation results are presented in~\cref{sec:results} and the paper is concluded in~\cref{sec:conclusion} by summarizing the conclusions and contributions, as well as discussing directions for future work.

\section{Vehicle Modeling}
\label{sec:vehicle_model}
This section presents the model of the tractor-trailer vehicle in the road-aligned frame. 
Additionally, it derives linear approximations of the position of the trailer axle that are suited for usage in numerical optimization.

\subsection{Road-aligned tractor-trailer model}
\begin{figure}
  \centering 
  \resizebox {0.99\columnwidth} {!} {
  \begin{tikzpicture}[scale=1]   
  \input{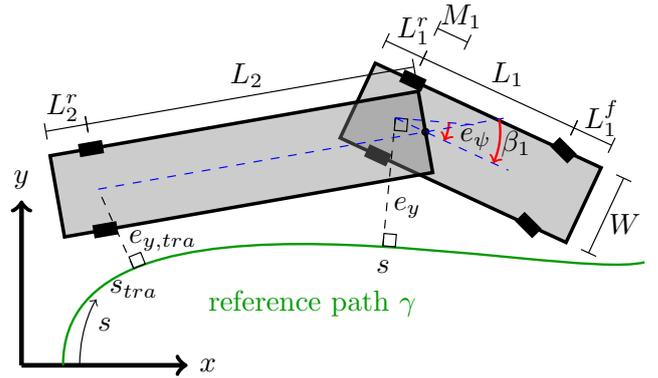}
  \end{tikzpicture}
  }
  \caption{{\FigureCaptionTextSize An illustration of the tractor-trailer vehicle in the road-aligned frame and definitions of relevant geometric lengths and vehicle states.}}
  \label{fig:road_aligned}
\end{figure}
The tractor-trailer vehicle considered in this work is composed of a car-like tractor and an interconnected trailer. The system is illustrated in~\cref{fig:road_aligned}.
The length $L_1$ corresponds to the tractor's wheelbase, length $L_2$ represents the distance between the center of the trailer's axle and the off-axle hitch connection at the tractor, and $M_1$ is the signed hitch offset (negative in~\cref{fig:road_aligned}).
Lengths $L_1^f$, $L_1^r$, and $L_2^r$ correspond to the overhangs of the vehicle, and $W$ denotes the width of the tractor and trailer.

In the road-aligned frame, the tractor-trailer vehicle is described in terms of deviation from a geometric reference path $\gamma$, as schematically illustrated in~\cref{fig:road_aligned}. Define $s$ as the distance traveled by the position of the tractor's rear axle onto its projection to the reference path $\gamma$, and let $\kappa_\gamma(s)$ be the curvature of the reference path. In this work, the reference path $\gamma$ corresponds to the center of the road lane, but could also correspond to the output solution of a global path planner. The tractor-trailer vehicle can be described by configuration vector $q=[s, e_y, e_\psi, \BetaTwo]^T$, where $e_y$ and $e_\psi$ represent the tractor's lateral and orientation error, respectively, with respect to the reference path. $\BetaTwo=\theta_\Veh-\theta_\Tra$ is the joint angle, defined as the difference between the orientation of the tractor $\theta_\Veh$ and the trailer $\theta_\Tra$. The evolution of the tractor states $(s, e_y, e_\psi)$ is equivalent to that of the road-aligned vehicle model used in~\cite{Gao:2012:Spatial}, whereas the model of the joint angle $\BetaTwo$ can be found in, e.g.,~\cite{Ljungqvist:2019:Path}. The model of the tractor-trailer vehicle in the road-aligned coordinate frame is given by
\begin{equation}
\label{eq:temporal_model}
\begin{aligned}
\dot{s} &= v\frac{\cos( e_\psi )}{1 - e_y\kappa_\gamma(s)}, \\
\dot{e}_y &= v\sin( e_\psi ), \\ 
\dot{e}_\psi &= v\left(\kappa - \frac{\kappa_\gamma(s)\cos( e_\psi )}{1-e_y\kappa_\gamma(s)}\right), \\ 
\dot{\BetaTwo} &= v \left( \kappa - \frac{\sin( \BetaTwo )}{\DTra} + \frac{\DHit}{\DTra} \cos( \BetaTwo )\kappa \right),
\end{aligned}
\end{equation}
where $\dot{q} = \nicefrac{\text{d} q}{\text{d}t}$ and $\kappa=\nicefrac{ \tan( \phi )}{ \DVeh }$ is the tractor's curvature.
The curvature of the tractor $\kappa$ is the control input, which is directly related to its steering angle $\phi$.
Since only forward motion is considered $v>0$, time-scaling can be applied to remove the time-dependency presented in~\eqref{eq:temporal_model} and transform the model into an equivalent spatial model~\citep{Gao:2012:Spatial}. Using the chain rule, it holds that $\nicefrac{\text dq}{\text d s }=\nicefrac{\text dq}{\text d t}\nicefrac{1}{\dot s}$, and the resulting spatial model becomes
\begin{equation}
\label{eq:spatial_model}
\begin{aligned}
e'_y &= \left(1 - e_y\kappa_\gamma\right)\tan( e_\psi ), \\ 
e'_\psi &= \frac{1 - e_y\kappa_\gamma}{\cos( e_\psi )}\kappa - \kappa_\gamma, \\ 
\BetaTwo' &= \frac{1 - e_y\kappa_\gamma}{\cos( e_\psi )}\left( \kappa- \frac{\sin( \BetaTwo )}{\DTra} +  \frac{\DHit}{\DTra}\cos( \BetaTwo)\kappa \right),
\end{aligned}
\end{equation}
where $(\cdot)'=\nicefrac{\text d(\cdot)}{\text ds}$ and $s'=1$. Therefore, the state vector is defined as $z = [e_y, e_{\psi}, \BetaTwo]^T$.
The spatial model in~\eqref{eq:spatial_model} is discretized and linearized as done in~\cite{Oliveira:2019:BusDriving}. 
First, the reference path $\gamma$ is discretized along its length resulting in $\{s_i\}^{N}_{i=0}$, with $s_i = i\DiscretizationSamplingDistance$, where $\DiscretizationSamplingDistance$ is the path sampling distance.
A linearization is then done around the reference states $\linearizationReferenceVariablesS = \{\bar{s}_{i}\}_{i=0}^{N}$, $\linearizationReferenceVariablesEy = \{\bar{e}_{y,i}\}_{i=0}^{N}$, $\linearizationReferenceVariablesEpsi = \{\bar{e}_{\psi,i}\}_{i=0}^{N}$, $\bm{\bar\beta}_1 = \{\bar{\beta}_{1,i}\}_{i=0}^N$ and $\bm{\bar\kappa} = \{\bar{\kappa}_{i}\}_{i=0}^{N}$, using a first-order Taylor approximation.
Thus, a linear discrete-time model in the form  $\VehState_{i+1} = A_i \VehState_{i} + B_i \kappa_i + G_i$ is obtained, where $\VehState_i = [e_{y,i}, e_{\psi,i}, {\beta}_{1,i}]^T$.

\subsection{Trailer-axle states}

The model in~\eqref{eq:spatial_model} describes behavior of the tractor's states as well as the joint-angle between the tractor and the trailer.
However, it does not contain direct information about the position and orientation of the trailer's axle.
For on-road planning purposes, it is of interest to also have equivalent trailer state variables $(\STrailer, \EYTrailer, \EPsiTrailer)$.
These state variables can then be used to define planning objectives, such as minimizing the lateral error of the trailer axle $\EYTrailer$, as seen later in \cref{subsec:optimization_objectives}.

Unfortunately, the road-aligned model does not allow for an analytical expression that relates the evolution of the trailer states $(\STrailer, \EYTrailer, \EPsiTrailer)$ as a function of the modeled ones $x$ and the tractor's curvature input $\kappa$. This is due to the distortions introduced in the road-aligned frame (discussed in detail in~\cite{Oliveira:2019:BusDriving, altafini:2002Following}).
The solution is to compute an approximate relationship of $(\STrailerApprox, \EYTrailerApprox, \EPsiTrailerApprox)$, which depends linearly on the tractor-trailer states in~\eqref{eq:spatial_model}.

Assuming a reference vehicle state $\bar x = [\SBar, \EYBar, \EPsiBar, \BetaTwoBar]^T$, we compute the corresponding position and orientation of the trailer's axle $(\STrailerBar, \EYTrailerBar, \EPsiTrailerBar)$ as:
\[
(\STrailerBar, \EYTrailerBar, \EPsiTrailerBar) = f( \SBar, \EYBar, \EPsiBar, \BetaTwoBar, \gamma ).
\]
Function $f$ first computes the equivalent Cartesian state of $\bar x$, it then computes the Cartesian position of the rear axle of trailer for that given state, and finally converts that position into the road aligned state $(\STrailerBar, \EYTrailerBar, \EPsiTrailerBar)$, by projecting it onto the reference path $\gamma$.
We note that function $f$ is not analytical due to the last step requiring the projection of a Cartesian position onto an arbitrary path $\gamma$ with varying curvature $\kappa_\gamma$. 
However, in the special case of a straight reference path, $f$ can be described by a closed-form expression~\citep{altafini:2002Following}.

To compute the approximation of $(\STrailerApprox, \EYTrailerApprox, \EPsiTrailerApprox)$ we then need to understand how $(\STrailerBar, \EYTrailerBar, \EPsiTrailerBar)$ changes with respect to states $(\EY, \EPsi, \BetaTwo)$.
This can be done by approximating the partial derivatives $\nicefrac{\partial \EYTrailer}{\partial \EY}$, $\nicefrac{\partial \EYTrailer}{\partial \EPsi}$, $\nicefrac{\partial \EYTrailer}{\partial \BetaTwo}$, $\nicefrac{\partial \EPsiTrailer}{\partial \EY}$, $\nicefrac{\partial \EPsiTrailer}{\partial \EPsi}$, and $\nicefrac{\partial \EPsiTrailer}{\partial \BetaTwo}$, e.g., using finite differences.
The linear approximation of the trailer states can then be defined as follows:
\begin{equation}
\label{eq:trailer_position_partial_derivatives_approx}
\begin{aligned}
\EYTrailerApprox =~&\EYTrailerBar + \frac{\partial \EYTrailer}{\partial \EY}(\EY - \EYBar)~+\\
& \frac{\partial \EYTrailer}{\partial \EPsi}(\EPsi - \EPsiBar) + \frac{\partial \EYTrailer}{\partial \BetaTwo}(\BetaTwo - \BetaTwoBar), \\
\EPsiTrailerApprox =~&\EPsiTrailerBar + \frac{\partial \EYTrailer}{\partial \EY}(\EY - \EYBar)~+\\ 
& \frac{\partial \EYTrailer}{\partial \EPsi}(\EPsi - \EPsiBar) + \frac{\partial \EYTrailer}{\partial \BetaTwo}(\BetaTwo - \BetaTwoBar).
\end{aligned}
\end{equation}
This model is a linear approximation of the lateral and orientation error of the trailer axle with respect to the reference path $\gamma$ at a fixed path length $\bar{s}$.

\section{Planning Approach}
\label{sec:problem_formulation}
In this section, the on-road path planning problem is formulated as an Optimal Control Problem (OCP) and an SQP approach to solve it is presented. Moreover, a set of different optimization objectives is presented targeting tractor-trailer on-road driving based on principles from human-like driving goals. 

\subsection{Optimal control problem}
The on-road path planning problem for the tractor-trailer vehicle is formulated as the following OCP:
\def\OptSpace{\quad}
\begin{subequations}
	\label{eq:optimization_problem}
\begin{align} 
    \underset{\bm \kappa}{\text{minimize}} & \OptSpace J_{e}^{k}( \VectorEY, \VectorEYTrailer ) + J_{\kappa}( \bm \kappa ) \label{eq:optimization_objective}\\
    \text{subject to} & \OptSpace \VehState_{i+1} = f( \VehState_{i}, \kappa_i ), \; i \in \{0, ..., N-1\}, \label{eq:constraint_1}\\
    & \OptSpace \VehState_0 = \VehState_{\mathrm{start}}, \; \kappa_0 = \kappa_{\mathrm{start}}, \label{eq:constraint_2}\\
    & \OptSpace \optConstraintObsPosition \leq g( \VehState_{i} ), \; i \in \{1, ..., N\}, \label{eq:constraint_3}\\
    & \OptSpace |\kappa_i| \leq \optConstraintMaxK, \; i \in \{1, ..., N-1\}, \label{eq:constraint_4}\\
    &\OptSpace |\kappa_i - \kappa_{i-1}| \leq \optConstraintMaxKDot,\; i \in \{1, ..., N-1\}, \label{eq:constraint_5}
\end{align} 
\end{subequations}
where $\VectorEY = [e_{y,1}~\allowbreak \ldots~\allowbreak e_{y,N}]^{T} \in \mathbb{R}^N$, $\VectorEYTrailer = [e_{y,\Tra,1}~\allowbreak \ldots~\allowbreak e_{y,\Tra,N}]^{T} \in \mathbb{R}^N$, and $\bm \kappa = [\kappa_{0}~\allowbreak \kappa_{1}~\allowbreak \ldots~\allowbreak \kappa_{N-1}]^{T} \in \mathbb{R}^N$.
The optimization objective~\eqref{eq:optimization_objective} is composed of two terms, the first term $J_{e}^{k}$ penalizes quantities related to the vehicle states, and the second term $J_\kappa$ that penalizes control inputs.
Different types of $J_{e}^{k}$ terms are discussed later in~\cref{subsec:optimization_objectives}.
In this work, the term $J_\kappa(\bm\kappa) = \sum_{i=1}^{N\text{-}1} \left( \kappa_i - \kappa_{i-1} \right)^2$ to enforce a smooth curvature profile which is directly related to a comfortable driving behavior.

The constraint in~\eqref{eq:constraint_1} corresponds to the vehicle model, whereas~\eqref{eq:constraint_2} defines the initial constraints on the vehicle states and control input.
Obstacle avoidance is ensured through constraint in~\eqref{eq:constraint_3}, which forces the vehicle's bodies to not collide with any obstacle nor exiting the road. For a more detailed definition of the constraints, the reader is referred to~\cite{Oliveira:2019:BusDriving}.
The last constraints in~\eqref{eq:constraint_4} and~\eqref{eq:constraint_5} define the tractor's curvature limitations including saturation $\optConstraintMaxK$ and rate limit $\optConstraintMaxKDot$.

To solve the OCP in~\eqref{eq:optimization_problem}, an SQP approach is used that is based on the work in~\cite{Oliveira:2019:BusDriving}. Between each SQP iteration, the vehicle model~\eqref{eq:constraint_1} is linearized around the solution of the previous iteration. Here, a first-order Taylor series approximation of the vehicle model~\eqref{eq:spatial_model} is used to obtain a linear prediction model of the tractor states $\VectorEY$ and $\VectorEPsi$ as well as the joint-angle state $\bm{\beta}_1$.
Furthermore, the trailer states $\VectorEYTrailer$ and $\VectorEPsiTrailer$ are obtained using the approximation given by~\eqref{eq:trailer_position_partial_derivatives_approx}.

\subsection{Optimization objectives}
\label{subsec:optimization_objectives}
In this section, we introduce a set of different candidate optimization objectives $J_{e}^{k}$ to be used in~\eqref{eq:optimization_problem}.

\subsubsection{Optimization objective 1 - Tractor centering}
In ordinary conditions, a vehicle is driving as much as possible in the center of its lane.
In this work, we assume that the reference path $\gamma$ of the road-aligned frame corresponds to the center of the lane.
Given such an assumption, centering the tractor is achieved by minimizing $|e_y|$ (note that if the tractor is driving precisely on the center, then $e_y=0$).
The first optimization objective $J_{e}^{1}$ is then defined to be the square of the euclidean norm of the lateral displacement of the tractor along the planned path:
\begin{equation*}
J_{e}^{1} = \left\| \VectorEY \right\|_{2}^{2}.
\end{equation*}

\subsubsection{Optimization objective 2 - Trailer centering}
In the case of articulated vehicles centering the tractor on the road might not suffice.
One should take into account the presence of the trailer, which might severely deviate from the center of the road, even when the tractor is centered.
This is notably true in turns, where the off-tracking effect is more critical, causing the trailer to significantly cut through the inside of the curve.
Thus, optimization objective $J_{e}^{2}$ is defined so as to minimize trailer displacement:
\begin{equation*}
J_{e}^{2} = \left\| \VectorEYTrailer \right\|_{2}^{2}.
\end{equation*}

\subsubsection{Optimization objective 3 - Tractor and trailer centering}
The third objective tries to center both the tractor and the trailer at the same time.
Instead of focusing either on the tractor or the trailer, we focus on minimizing the lateral error of both.
This is achieved by defining the third optimization objective as:
\begin{equation*}
 J_{e}^{3} = \left\| (1-K)\VectorEY + K\VectorEYTrailer \right\|_{2}^{2},
\end{equation*} 
where $K \in [0, 1]$ is a design parameter. The value of $K$ specifies the important trade-off between centering the tractor or the trailer around the road center. A method for determining a suitable $K$ will later be presented in~\cref{subsec:different_optimization_objectives}.

\subsubsection{Optimization objective 4 - Tractor and trailer maximum deviation minimization}
The fourth candidate, unlike the previous ones, uses the $\mathcal L_\infty$-norm.
By using an $\mathcal L_\infty$-norm, we minimize the worst lateral deviation of the vehicle states.
Intuitively, we expect this to result in planned paths that minimize the maximum lateral deviation of the vehicle axles from the road center.
The fourth optimization objective is then defined as:
\begin{equation*}
J_{e}^{4} = \| (\VectorEY, \VectorEYTrailer) \|_{\infty}.
\end{equation*}
\subsubsection{Optimization objective 5 - Swept area minimization}
\begin{figure}
  \centering 
  \resizebox {0.99\columnwidth} {!} {
  \begin{tikzpicture}[scale=0.95]   
  \input{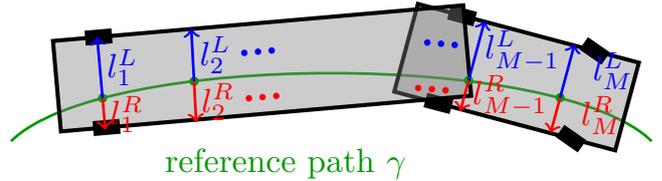}
  \end{tikzpicture}
  }
  \caption{
  {\FigureCaptionTextSize Illustration of the auxiliary variables $l_{1}^{L}, \ldots, l_{M}^{L}, l_{1}^{R}, \ldots, l_{M}^{R}$ that are sampled uniformly across the vehicle length and used to measure the lateral displacement of the vehicle sides. These variables provide an estimate of how much the vehicle sides deviate from the center of the road.}}
\label{fig:minimization_of_sides}
\end{figure}

The last objective focuses on minimizing the distance of the vehicle sides to the center of the road.
To do so, we define the vector of auxiliary variables $\mathbf{q} = [l_{1}^{L}, l_{2}^{L}, \ldots, l_{M}^{L}, l_{1}^{R}, l_{2}^{R}, \ldots, l_{M}^{R}]^T$, as shown in~\cref{fig:minimization_of_sides}.
The superscripts $L$ and $R$ correspond to the left (L) and right (R) sides of the vehicle.
Vector $\mathbf{q}$ measures the displacement of the vehicle's sides to the center of the road.
A vector $\mathbf{q}_i$ is computed for each vehicle state $(e_{y,i}, e_{\psi,i}, \beta_{1,i})$.
In the ideal case, one would like to keep the vector $\mathbf{q}_i$ as small as possible, resulting in the tractor-trailer vehicle body driving as close as possible to the center of the road.
Thus, the optimization objective $J_{e}^{5}$ is defined as:
\begin{equation*}
J_{e}^{5} = || ( \mathbf{q}_1, \mathbf{q}_2, \ldots, \mathbf{q}_N ) ||_{\infty}.
\end{equation*}
With this optimization objective, we expect the path planner to find a solution that minimizes the most significant displacement of the vehicle sides to the center of the road.

Note that this objective directly uses the positional information of the vehicle sides in its formulation, and therefore takes into account the vehicle's dimensions.
This is unlike the previous candidate objectives that only make use of the axle positions to center the tractor-trailer vehicle.

\section{Simulation Results}

\def\EnvelopesComparisonFilename{figures/script_08_c_c2_vehicle_envelop_road.tikz}
\def\SelectedMetricPathPlanned{figures/script_08_c_d_show_metric_3_K_0_4.tikz}
\def\RoundaboutResult{figures/script_07_a_roundabout_cebon.tikz}
\def\ObstacleAvoidanceResult{figures/script_04_b_obstacle_avoidance_u_turn.tikz}

\label{sec:results}
In this section, the performance of the proposed on-road path planner with the different optimization objectives is evaluated using a set of relevant performance metrics. Furthermore, simulation results for two realistic urban driving scenarios are presented to highlight the capabilities of the proposed path planner. We use the vehicle parameters and dimensions for the tractor-trailer vehicle shown in~\cref{fig:truck-trailer-photo} which are summarized in Table~\ref{tab:vehicle_parameters}. Note that the total length of the tractor-trailer vehicle sums up to $24$ meters, which corresponds to the maximum length legally allowed in Sweden, and therefore one of the most challenging tractor-trailer dimensions that are allowed to drive on public roads in Sweden. The simulations have been performed on a laptop computer with an Intel Core i7‐6820 HQ@2.7GHz CPU. The path planner has been implemented in MATLAB where CVX is used as convex solver~\citep{Grant:2014:cvx} in each SQP iteration.

\begin{table}[t!]
	\caption{{\FigureCaptionTextSize Vehicle parameters of the tractor-trailer vehicle.}}
	\centering
	\begin{tabular}{|l| l|}
		\hline 
		Vehicle parameter  & Value   \\  \hline 
		Tractor's wheelbase $L_1$            &   $3.78$ m  \\ 
		Tractor's rear overhang $L_1^r$      &   $1.64$ m \\
		Tractor's front overhang $L_1^f$     &   $1.46$ m \\
		Length of off-hitch $M_1$            &   $-0.30$ m  \\
		Width of tractor and trailer $W$     &   $2.54$ m \\
		Length of trailer $L_2$              &   $13.97$ m  \\ 
		Trailer's rear overhang $L_2^r$      &   $4.50$ m \\
		Tractor's maximum curvature $\kappa_{\text{max}}$ & $0.1~\text{m}^{-1}$ \\   
		Tractor's curvature-rate limit $\optConstraintMaxKDot$ & 0.1$\Delta s$ \\
		\hline 
	\end{tabular}
	\label{tab:vehicle_parameters}
\end{table}

\subsection{Performance metrics}
To understand the performance of the different optimization objectives proposed in Section~\ref{subsec:optimization_objectives}, a set of performance metrics are introduced.
First, we would like to measure the maximum amount the vehicle bodies sweep on the road.
Performance metrics \emph{max left} and \emph{max right} measure the maximum offset from any point on the vehicle body to the center of the road.
\emph{max left} measures the maximum amount to the left of the road center, whereas \emph{max right} measures the amount to the right of the road center.
An illustration of these metrics can be seen in~\cref{fig:performed_path_metric_3}.

The second metric \emph{$a_L - a_R$} measures the difference between the areas swept by the vehicle bodies to the left and to the right of the road center.
In the case of a straight road, and if the vehicle drives exactly on the center, the value of this metric would be zero.
Large values of this metric indicate a preference for the vehicle to be off-centered, with a positive value indicating a tendency to drive on the left of the road center, and a negative value indicating a tendency to drive on the right.
The closer this metric is to zero, the more centered the tractor-trailer drives.

The final metric \emph{CPU time} measures the amount of time required to solve the OCP in~\eqref{eq:optimization_objective} using different optimization objectives. This metric is used to compare the computational effort that the different optimization objectives require from a computing unit.

\subsection{Comparison of different optimization objectives}
\label{subsec:different_optimization_objectives}

\begin{figure}
  \centering
     \resizebox {0.99\columnwidth} {!} { \input{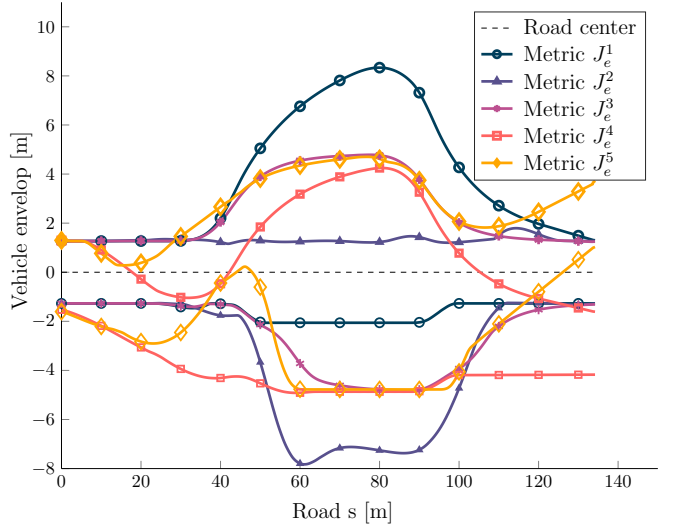} }
  \caption{{\FigureCaptionTextSize Comparison of the maximum offset and swept area by the articulated vehicle using different optimization objectives. \Cref{tab:swept_area_results} provides detailed information about the swept areas and maximum offsets.}}
  \label{fig:comparison_sweeped_paths}
\end{figure}

\def\Veh{\text{veh}}
\def\Tra{\text{tra}}
\def\VehTra{\text{veh,tra}}
Here, we compare the results of using the different optimization objectives introduced in~\cref{subsec:optimization_objectives}.
\cref{fig:comparison_sweeped_paths} shows the envelopes of the areas swept by the vehicle when performing a U-turn, and using different objectives.
\Cref{tab:swept_area_results} presents in detail performance measurements of each of the objectives for a selected U-turn with a curvature of $0.065~\text{m}^{-1}$ (turning radius of $15.38$ m).
The optimization has a planning horizon of $134.2$ m, and a discretization of the path of $0.1$ m.
An individual study of each optimization objective and its performance follows.

\def\tikzLegendSize{ \large }

The first objective, $J_{e}^{1}$, results in the largest positive (left) sweep, corresponding to the trailer taking the turn on the inside up to $8.34$ m.
Furthermore, the vehicle has a swept area, defined as the difference between the area swept to the left and right of the lane center, of $312~\text{m}^2$, indicating that it tends to the inside of the turn.
This is a direct result of the objective formulation, which tries to keep the tractor on the road center while disregarding the trailer.
Thus, the trailer cuts too much on the inside of curves, making this optimization objective unsuitable.

Likewise, the second objective $J_{e}^{2}$ results in the largest negative (right) sweep, corresponding to the tractor taking the turn on the outside up to $7.83$ m from the road center.
The swept area is $-302~\text{m}^2$, indicating a clear preference for driving on the outside of the turn.
This is not surprising, as the formulation only attempts to keep the trailer rear axle centered on the road.
In order to keep the trailer on the road center, the tractor drives excessively on the outside, making this optimization objective unsuitable.

Before evaluating objective function $J_{e}^3$, we first need to find the best value $K$.
To do so, we perform a discrete search over the interval $K \in [0, 1]$, and measure the performance of the resulting planned path with respect to the area difference metric \emph{$a_L - a_R$}.
We run the path planner with different values of $K$ for several U-turn roads of different curvatures and obtain that $K = 0.45$ is consistently performing the best for all roads.
Therefore, we select $K = 0.45$ in optimization objective $J_{e}^{3}$.
Using the $K$ in optimization objective $J_{e}^{3}$, results in a balanced trade-off between the cut-in of the trailer and the cut-out of the tractor, as well as a small swept area of $-5~\text{m}^2$.

Optimization objective $J_{e}^{4}$ achieves a better trade-off between the cut-in and cut-out than $J_{e}^{1}$ and $J_{e}^{2}$, however it is significantly worse than $J_{e}^{3}$.
Furthermore, since this objective uses the infinity norm, the vehicle does not have any incentive to come back to the center of the road after the turn finishes, as shown in~\cref{fig:comparison_sweeped_paths}.
This happens because the infinity norm in $J_{e}^{4}$ only penalizes the vehicle state that leaves the road center the most, resulting in an excessively large swept area of $-406~\text{m}^2$.

Finally, optimization objective $J_{e}^{5}$ shows the a quite good trade-off between cutting in and out of the road, resulting in a good balance between maximum cut-in and maximum cut-out.
We note that this trade-off is slightly worse than the one achieved by $J_{e}^{3}$, however it does not rely on the tuning of parameter $K$ as is the case with $J_{e}^{3}$.
Similarly to $J_{e}^{4}$, objective $J_{e}^{5}$ also suffers from the problems associated with the infinity norm.
We can see in~\cref{fig:comparison_sweeped_paths} that the vehicle does not converge to the center of the road at the final section of the path.
Moreover, we note that objective $J_{e}^{5}$ makes the optimization problem very expensive to solve, as a CPU time of $770.71$ s is required to solve the OCP.
This optimization objective is therefore not suitable for implementation on a real system with online planning requirements.
\def\tikzLegendSize{ \large }
\begin{figure}[t!]
	\centering
	\resizebox {0.9\columnwidth} {!} { \input{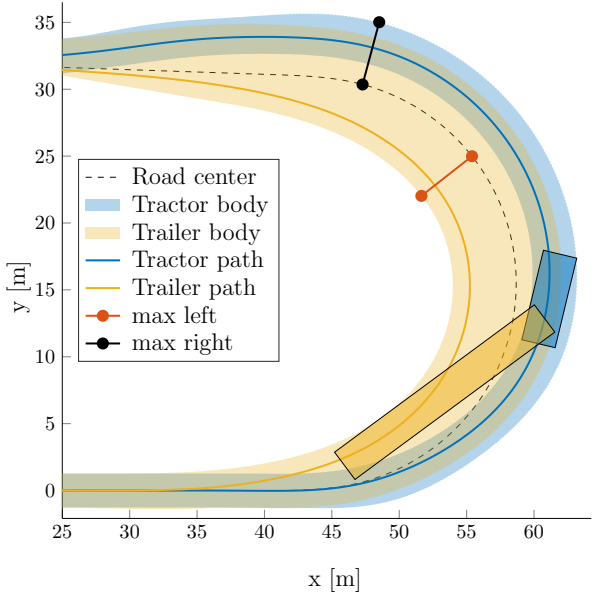} }
	\caption{{\FigureCaptionTextSize Path performed by the tractor-trailer when using optimization objective $J_{e}^{3}$. The planned solution fairly balances the maximum cut-out of the tractor to the right of the road center, and the maximum cut-in of trailer to the left of the road center. \emph{max left} and \emph{max right} measure the maximum amount of sweep of the vehicle body to the left and right of the road center.} }
	 \label{fig:performed_path_metric_3}
\end{figure}

We conclude that optimization objective $J_{e}^{3}$ is the most suited for our purposes.
Firstly, it achieves a low swept area, as well as a balanced trade-off between maximum cut-in and cut-out.
Secondly, since it penalizes all states along the planned path, it takes the vehicle to the center of the road in both turning and straight segments.
Thirdly, its CPU time of $22.03$ s makes it promising for usage in online planning for autonomous vehicles.

\begin{remark}
The previous analysis focuses on a specific U-turn defined by a curvature of $0.065~\text{m}^{-1}$.
The same analysis was also made for several other U-turns with different curvatures, as well with curvatures with different sign (turning clockwise, as opposed to counterclockwise).
The comparisons and conclusions made previously are also observed in the majority of distinct U-turns we have tested.
\end{remark}

\begin{remark}
The CPU times were measured in a MATLAB implementation running on a personal laptop, and the optimization problem was solved for the whole length of the road.
We expect that a real implementation could greatly speed up the computation times, due to: 1) implementation in a low-level language, such as \texttt{C++}; 2) reformulation of the QP problem and use of a tailored solver; 3) execution in a receding horizon fashion, which will significantly reduce the length of the road considered for planning (and therefore the number of vehicle states to be optimized), and provide initial guesses to the optimization problem allowing for efficient warm-starts.
\end{remark}

The planned path for the tractor-trailer vehicle using optimization objective $J_{e}^{3}$ is shown in~\cref{fig:performed_path_metric_3}.
We observe that the optimal solution achieves a balanced trade-off between the cut-in of the trailer body and the cut-out of the tractor body.

\def\TableFontSize{\small}
\begin{table}
\centering
\caption{{\FigureCaptionTextSize Performance indexes of the different optimization objectives. The measures \emph{max left} and \emph{max right} correspond to the maximum amount swept by the vehicle body to the left and right of the road center, respectively. \emph{$a_L - a_R$} is the difference between the area swept to the left (L) and to the right (R) of the road center. \emph{CPU time} corresponds to the time required to solve the OCP in~\eqref{eq:optimization_problem} using the different optimization objectives.}}
\begin{tabular}{|c|c|c|c|c|}
\hline
{\TableFontSize Objective} & {\TableFontSize max left} & {\TableFontSize max right} & {\TableFontSize $a_L - a_R$} & {\TableFontSize CPU time} \\
\hline
$J_{e}^{1}$ & 8.34 m & $2.06$ m & $312~\text{m}^2$ & 7.92 s \\
\hline
$J_{e}^{2}$ & 1.79 m & $7.83$ m & $-302~\text{m}^2$ & 17.93 s \\
\hline
$J_{e}^{3}$ & 4.79 m & $-4.82$ m & $-5~\text{m}^2$ & 22.03 s \\
\hline
$J_{e}^{4}$ & 4.25 m & $-4.82$ m & $-406~\text{m}^2$ & 13.53 s \\
\hline
$J_{e}^{5}$ & 4.70 m & $4.78$ m & $4~\text{m}^2$ & 770.71 s \\
\hline
\end{tabular} 
\label{tab:swept_area_results} 
\end{table}

\def\tikzLegendSize{ \large }
\begin{figure}[t!]
	\centering
	\resizebox {\columnwidth} {!} { \input{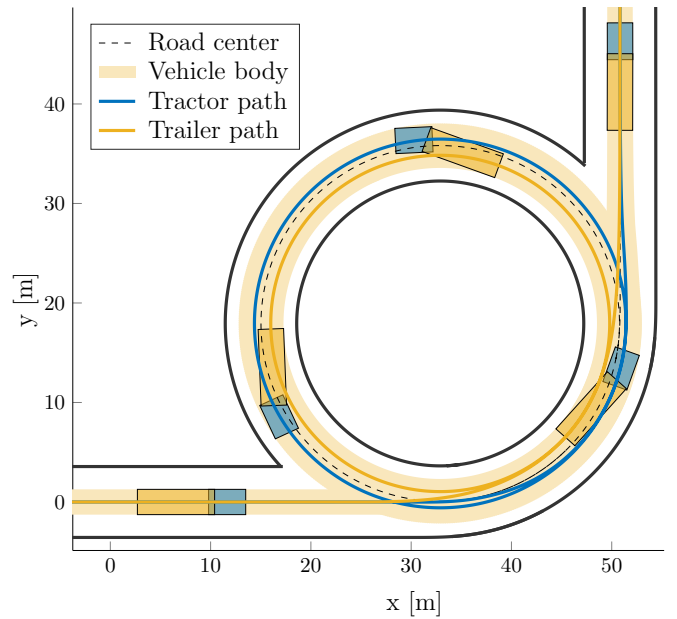} }
	\caption{{\FigureCaptionTextSize The tractor-trailer vehicle performing a 450-degree turn in a roundabout. The planned path successfully balances the tractor and trailer cut-out and cut-in towards the right and left sides of the road center. The roundabout and vehicle dimensions are based on the work in~\cite{Jujnovich:2013:SteeringControl}.}}
	\label{fig:driving_on_roundabout} 
\end{figure}

\subsection{Driving on a roundabout}

In this section, the performance of the proposed path planner using optimization objective $J_{e}^{3}$ is evaluated on a 450-degree turn performed in a roundabout that is illustrated in~\cref{fig:driving_on_roundabout}.
The roundabout scenario is equivalent to the one presented~\cite{Jujnovich:2013:SteeringControl}, where it is used to test for compliance with UK requirements for roundabout maneuvers. The curvature of the roundabout is $0.056~\text{m}^{-1}$ (turning radius of $17.88$ m), the planning horizon of the optimization problem is $245.8$ m and the sampling distance is $0.2$ m. Moreover, the work in~\cite{Jujnovich:2013:SteeringControl} considers a smaller tractor-trailer combination, which we mimic by setting the tractor-trailer vehicle dimensions to the ones listed in~\cref{tab:vehicle_parameters_cebon}.

\begin{table}[t!]
	\caption{{\FigureCaptionTextSize Vehicle parameters for the tractor-trailer vehicle used in the roundabout scenario. Values partially based on~\cite{Jujnovich:2013:SteeringControl}.}}
	\centering
	\begin{tabular}{|l| l|}
		\hline 
		Vehicle parameter  & Value   \\  \hline 
		Tractor's wheelbase $L_1$            &   $3.47$ m  \\
		Tractor's rear overhang $L_1^r$      &   $1.34$ m \\
		Tractor's front overhang $L_1^f$     &   $1.16$ m \\
		Length of off-hitch $M_1$            &   $-0.30$ m  \\
		Width of tractor and trailer $W$     &   $2.54$ m \\
		Length of trailer $L_2$              &   $9.40$ m  \\
		Trailer's rear overhang $L_2^r$      &   $3.03$ m \\
		Tractor's maximum curvature $\kappa_{\text{max}}$ & $0.1~\text{m}^{-1}$ \\   
		Tractor's curvature-rate limit $\optConstraintMaxKDot$ & 0.1$\Delta s$ \\
		\hline 
	\end{tabular}
	\label{tab:vehicle_parameters_cebon}
\end{table}

We note that since we are considering a vehicle with significantly different dimensions, we need to adjust parameter $K$ of optimization objective $J_{e}^{3}$.
To do so, we run the path planner with different values of $K$ in several U-turn roads, and select $K = 0.40$ as it is the parameter that performs best for all roads.

The simulation results are presented in~\cref{fig:driving_on_roundabout}. The optimal solution takes $11.88~s$ to compute and as can be seen, the planned path properly balances the cut-out and cut-in of the tractor and trailer as it drives along the roundabout.
The vehicle smoothly enters and exits the roundabout, while keeping its swept width reasonably small at all times.

\subsection{Collision avoidance}
We set up a challenging scenario to illustrate the collision avoidance capabilities, enforced via constraints in~\eqref{eq:constraint_3}.
We again consider a vehicle with the maximum allowed dimensions, as reported in Table~\ref{tab:vehicle_parameters}.
In this scenario, the vehicle drives along a sharp U-turn, with a curvature of $0.040~\text{m}^{-1}$ (turning radius of $25$ m).
Two obstacles are placed on the road, resulting in the planned solution shown in~\cref{{fig:collision_avoidance_u_turn}}.
We see the vehicle initially turning towards the outside of the turn to avoid a collision with the first obstacle.
The obstacle avoidance constraints of the trailer body dictate this maneuver, guiding the tractor towards the outside so that the trailer safely avoids the obstacle.

Later on, the second obstacle forces the vehicle to the inside of the turn.
This maneuver occurs due to the obstacle avoidance constraints of the tractor body, that force the vehicle to avoid the obstacle.
The trailer follows safely, as it is dragged along through the inside of the turn.

\def\tikzLegendSize{ \normalsize }
\begin{figure}
  \centering
	\resizebox {\columnwidth} {!} { \input{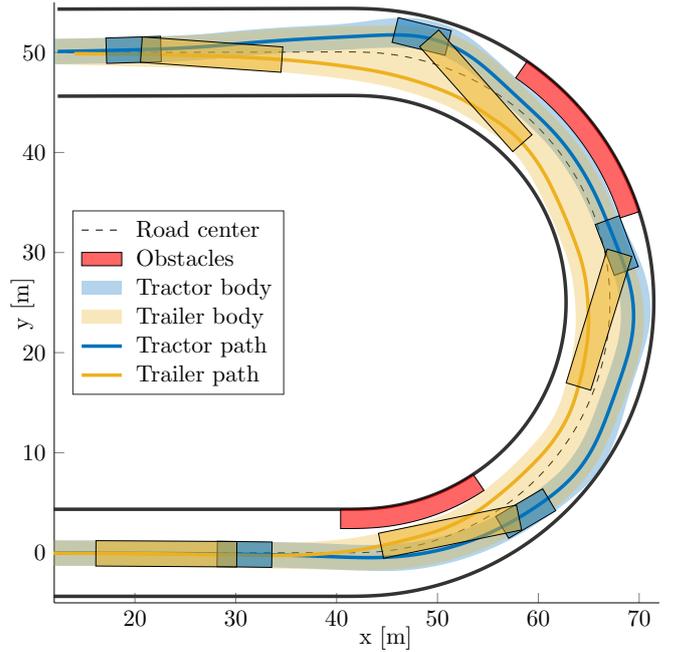} }
  \caption{{\FigureCaptionTextSize The tractor-trailer vehicle avoiding obstacles located in a U-turn. The first maneuver swerves the vehicle so that the trailer avoids the obstacle on the inside of the turn. The second maneuver takes the tractor to the inside of the turn, avoiding a collision with the second obstacle. Both tractor and trailer converge to the road center after the turn.}}
\label{fig:collision_avoidance_u_turn}
\end{figure}

In this scenario, the curvature of the U-turn is $0.065~\text{m}^{-1}$ (turning radius of $15.38$ m), the planning horizon of the optimization problem is $134.2$ m with a sampling distance of $0.2$ m, and the measured computation time is $108.41$ s.
We comment that such a high computation time is due to the complexity of the maneuver, requiring the SQP solver to perform several iterations until converging to a feasible solution.
The dimensions of the vehicle severely impact computation times, as noticed by the fact that the same scenario solved for a vehicle with the dimensions used in~\cite{Jujnovich:2013:SteeringControl} results in a computation time of $29.89$ s, significantly lower than the time required for the case of the larger vehicle.

\subsection{Model fidelity}

The road-aligned tractor-trailer model is based on the well studied kinematic bicycle model.
\cite{Kong:2015:KinematicDynamicModels} shows that this model is suitable for describing vehicle movement at low lateral forces, which corresponds to the use case of the articulated vehicles studied in this work.

The linearization and discretization of the vehicle model introduce errors in the vehicle model.
However the usage of an SQP strategy helps to address these issues.
By sequentially linearizing the problem and solving until convergence, the SQP ensures that the planned solution path is arbitrarily close to the linearization reference.
This convergence ensures that the planned path follows the nonlinear kinematic model.

Moreover, in practical applications, the proposed planner would be implemented in a receding horizon fashion.
In this way, the optimal planned solution computed is only used during the following planning interval.
At the next planning interval, the planner computes a new path based on the current vehicle state and environment observations over the shifted horizon.
Thus, the planner works in a closed-loop, alleviating possible modeling errors that might arise.

\section{Conclusion}

\label{sec:conclusion}
We have proposed an optimization-based on-road path planner for tractor-trailer combinations, as well as articulated buses, driving in urban environments.
The planner solves an optimal control problem using a tractor-trailer road-aligned vehicle model and an iterative method for computing the off-tracking, as well as approximate partial derivatives of each point of the vehicle bodies. 
Furthermore, we propose and study in detail a set of candidate optimization objectives for the optimal control problem, showing that standard passenger vehicle objectives are not suitable for tractor-trailer vehicles. 
We then select the optimization objective that renders in optimized paths which result in less intrusive driving caused by the swept path of the tractor-trailer vehicle's bodies.
Finally, we present two challenging urban scenarios, a U-turn with obstacles, and a 450-degree roundabout.
In both scenarios, the proposed method plans a human-like path while ensuring collision avoidance and centering of the tractor and trailer bodies.
Computation times show that the proposed planner has promising capabilities of being further developed into an algorithm to be implemented in real vehicles. 

As future work, we plan to make the solution suitable for implementation in real vehicle hardware. Thus, it becomes necessary to reduce computation times, which we expect to achieve by using a low-level programming language, a receding-horizon approach to the optimal control problem, and a tailored QP solver.
Besides, we would also like to consider more complicated combinations of articulated vehicles, which can either be composed of more vehicle bodies, or actively-steered trailer wheels.

\bibliography{ifacconf}             

\end{document}